\def\year{2020}\relax
\documentclass[letterpaper]{article} 
\usepackage{aaai20}  
\usepackage{times}  
\usepackage{helvet} 
\usepackage{courier}  
\usepackage[hyphens]{url}  
\usepackage{graphicx} 
\usepackage{graphicx}  
\title{One-Shot Image Classification by Learning to Restore Prototypes}
\author{Wanqi Xue\\
Department of Computer Science\\
National University of Singapore\\
wanqixue0@gmail.com\\
\And
Wei Wang\\
Department of Computer Science\\
National University of Singapore\\
wangwei@comp.nus.edu.sg\\
}

\newcommand{\eg}{e.g. }
\newcommand{\etal}{et al. }
\usepackage{bm}

\begin{document}

\maketitle

\begin{abstract}
One-shot image classification aims to train image classifiers over the dataset with only one image per category. It is challenging for modern deep neural networks that typically require hundreds or thousands of images per class. In this paper, we adopt metric learning for this problem, which has been applied for few- and many-shot image classification by comparing the distance between the test image and the center of each class in the feature space. However, for one-shot learning, the existing metric learning approaches would suffer poor performance because the single training image may not be representative of the class. For example, if the image is far away from the class center in the feature space, the metric-learning based algorithms are unlikely to make correct predictions for the test images because the decision boundary is shifted by this noisy image. To address this issue, we propose a simple yet effective regression model, denoted by RestoreNet, which learns a class agnostic transformation on the image feature to move the image closer to the class center in the feature space. Experiments demonstrate that RestoreNet obtains superior performance over the state-of-the-art methods on a broad range of datasets. Moreover, RestoreNet can be easily combined with other methods to achieve further improvement. 
\end{abstract}

\section{Introduction}
Over the past decade, we have witnessed the great success of deep learning in computer vision. With large amounts of annotated data, deep learning models achieved impressive breakthroughs again and again~\cite{resnet,alexnet,densenet}. However, in practical applications, large quantities of labeled data is expensive or sometimes impossible to acquire. In such a situation where only a few samples per category are available, both training from scratch and fine-tuning on the small dataset are likely to cause severe overfitting, leading to poor recognition performance. Humans, in contrast, have the ability to quickly learn a new concept from one or a few examples. The significant gap between human and machine intelligence encourages the interest of researchers. Many endeavours have been done to narrow the gap~\cite{maml,protonet,matchingnet,metalearnlstm,metanetworks}. 
\begin{figure}
\centering
    \begin{center}
        \includegraphics[width=0.35\textwidth]{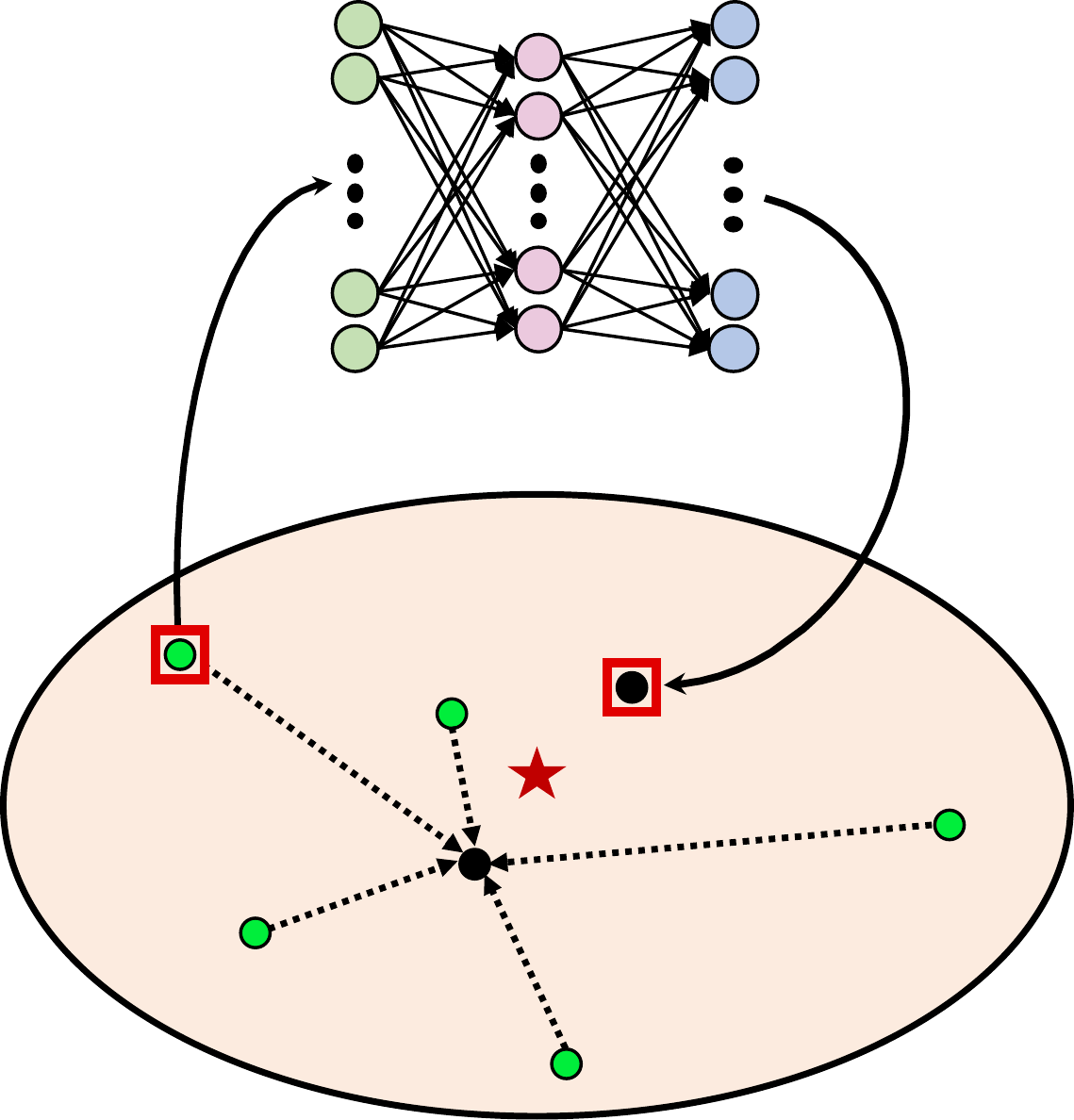}
    \end{center}
    \caption{The challenge for metric learning in one-shot image classification and our solution to it. Each green point represents a training image. Dark points denote the prototypes and the red star marks the center of the class.}
    \label{sketch}
\end{figure}

\begin{figure*}
\centering
\includegraphics[width=0.9\textwidth]{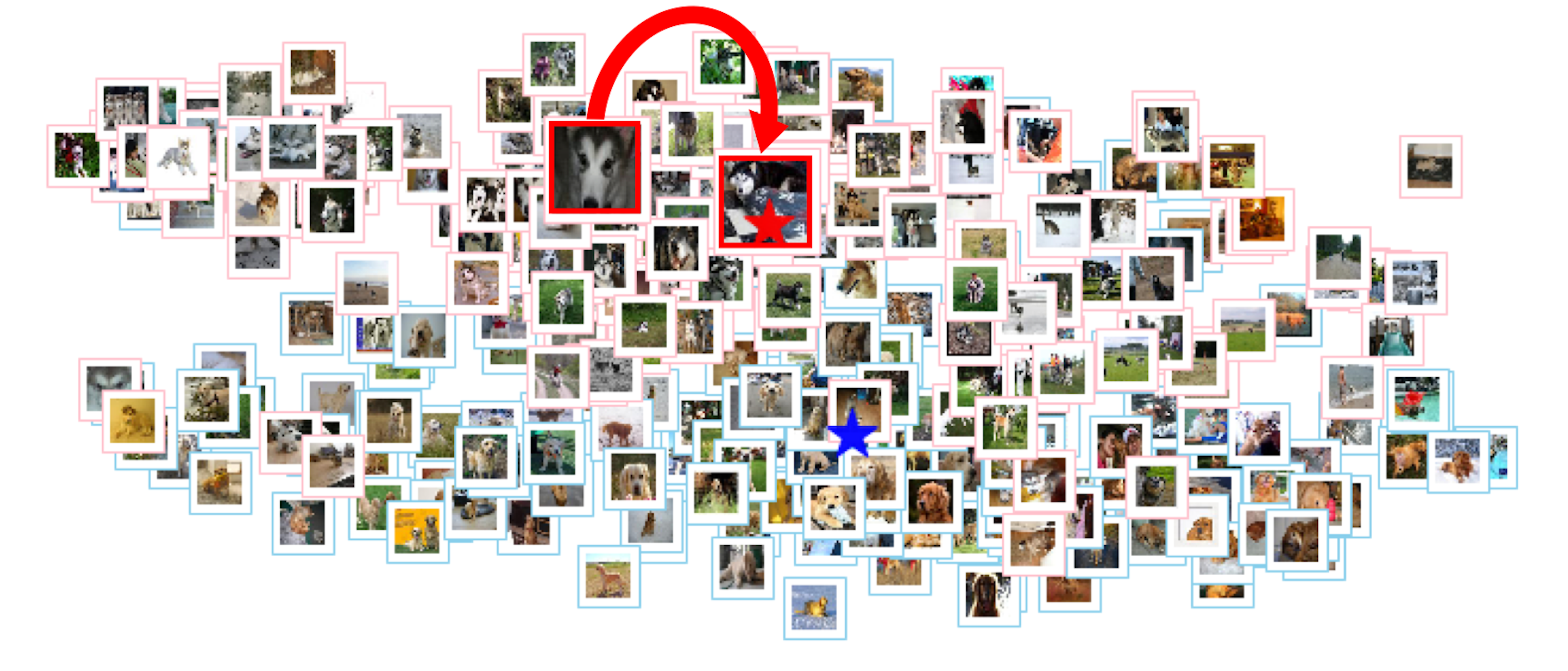}
\caption{2D visualization of image features for one-shot classification of two classes, namely Alaskan Malamute (framed in pink) and Golden Retriever (framed in blue). After restoration, the original prototype of Malamute (framed in red, left) is moved to the right, which is closer to the class center (marked by the red star). We visualize the shifted prototype (right) using its nearest real image. The figure is plotted by applying t-SNE of ResNet18 features of samples from  \textit{mini}ImageNet. Best viewed in color.}
\label{first_figure}

\end{figure*}
A popular category of solutions is based on meta-learning, where a meta-learner is trained to generate classifiers. The training is conducted in episodic manner, where the each episode is constituted by two sets of data, namely the support set and the query set. The generated classifier is trained over the support set and evaluated on the query set. The meta learner is then updated based on the evaluation performance. The idea is to transfer some class agnostic knowledge from the training data to the test data via the meta learner. Different meta-learning approaches have been proposed. Metric-learning-based methods, \eg Prototypical network~\cite{protonet}, train the meta-leaner to transform the images into a metric space where nearest neighbour classifiers can be applied. MAML trains the meta-leaner to learn a good initialization state~\cite{maml,metasgd} for the convolutional neural network (ConvNet) classifiers.

Metric-learning-based approaches are simple and effective. However, they would suffer from poor performance for one-shot learning, i.e., learning from only one single example per category. Take Figure \ref{sketch} as an example, when there are more training images, the average of these images' features, denoted as the prototype, is more likely to be around the real class center, even though some images are far away from the center. In contrast, if there is only one single training image and it is far away from the center, then the nearest neighbour classifier is unlikely to make correct predictions for the test images as the decision boundary is shifted by this noisy image (i.e., the prototype). 

In this paper, we focus on one-shot learning and propose a simple solution towards the issue mentioned above. The intuition of our solution is to train a transformation network in the feature space to move the noisy training image close to the center of the cluster.
During training, \emph{RestoreNet} learns from training pairs constituted by the features of noisy images and their corresponding class prototypes. Each class prototype is constructed using many images from the class and thus is reliable. During test, the feature of the image from the support set is fed into \emph{RestoreNet}. We average the original feature and the transformed feature to get the final image representation, which is used as the prototype of its class for the classification of the query images (via nearest neighbour classification). Figure \ref{first_figure} presents the  visualization of the image features for one-shot classification of two classes, namely Alaskan Malamute and Golden Retriever. 
The example training image of Alaskan Malamute (framed in red, left) merely includes the head and its representation (i.e., the prototype) is located far from the center of its cluster (marked by red star). After restoration, the original prototype is moved to the right, which is closer to the class center.

Our contribution is three-fold: firstly, we identify the challenge of metric-learning-based approaches for one-shot learning. Secondly, we propose a simple method, i.e., \emph{RestoreNet}, to address the challenge by moving the generated class prototype closer to the class center in the feature space. The proposed model can be combined with other methods easily and realize further enhancement. Finally, experiments on four benchmark datasets demonstrate that our model improves significantly over the state-of-the-art methods for one-shot learning tasks.

\section{Related Work}
One-shot (resp. few-shot) learning requires the classifiers to quickly adapt to new classes using only one (resp. few) example from each target class. Fine-tuning classifier on such sparse data is likely to get severe overfitting. To address this problem, different approaches have been proposed.

\textbf{Data augmentation} \cite{deltaencoder,semanticaug,wyxendtoend,fairiccv} resolve the data issue directly by data augmentation. Delta-encoder~\cite{deltaencoder} applies the extracted intra-class deformations or ``deltas'' to the one-shot (resp. few-shot) example of a novel class to generate new samples. \cite{wyxendtoend,semanticaug} trains a network which can effectively mix noise with image representations. They generate data by applying different noise.
Another way to increase the training dataset is via self-training~\cite{self}. In self-training framework, a predictor is first learned on the initial training set. Then the predictor is applied to predict the labels of a set of unlabeled images. These images are added to the training set to re-train the original predictor. 
In \cite{semi}, the unlabeled images are assigned with weights before they are added to the training set, which are used to sample the images to create the training batches. In \cite{selfjig}, images from the original training set and the unlabeled images are edited to synthesize new images. Our proposed solution is orthogonal to these data augmentation methods. To confirm it, we adopt the first approach in one of our experiments.

\textbf{Meta-learning} Many recent works~\cite{memorynet,metanetworks,relationnet,matchingnet,protonet,modelregression,modeltail} follow the meta-learning paradigm. They train a meta-learner over a series of training episodes. The meta-learner then generates the classifier for the target task by exploiting the accumulated class agnostic knowledge. For example, \cite{maml,metasgd,metalearnlstm} train the meta-learner to find a good initialization state and/or learn an effective optimizer, which are applied to optimize the classifier for the target task. Then the classifier can generalize better to new data within a few gradient-descent update steps. Metric learning~\cite{relationnet,matchingnet,protonet} based approaches can also be formalized under the meta-learning paradigm. They aim to learn a feature space where the prototype of each class is the center of the training images in the class. Nearest neighbour classification is applied to classify the query images using Euclidean distance, cosine distance, etc. Recently, \cite{modelregression,modeltail} propose to train a meta-learner to transform the classifier trained over few examples to the classifier trained over many examples by adapting the classifier parameters. This kind of approaches are somewhat similar to $\textit{RestoreNet}$. However, there are mainly two differences: 1) the motivation. We try to improve the performance by adjusting the FEATURE of NOISY examples. Therefore, we train $\textit{RestoreNet}$ using the farthest example, which are considered as noisy examples. \cite{modelregression} tries to improve the performance by adjusting the few-shot CLASSIFIER (parameters) to be similar to the many-shot classifier where the transform model is trained using RANDOMLY selected few-shot classifiers. 2) consequently, different techniques are applied. We get the FEATURE of each example by averaging (like ensembling) the original feature and transformed feature, whereas \cite{modelregression} uses the transformed model as biased regularization.

\section{Methodology}
\subsection{Background}
\textbf{Problem Definition} 
For N-way K-shot learning, we are given a support set of labeled images $\mathcal{S}_{novel}=\{(\bm{x_i}, y_i)\}_{i=1}^{N*K}$, where $\bm{x_i}$ is the image, $y_i\in C_{novel}$ is the label, $C_{novel}$ is the class set, $N$ is the number of classes in $\mathcal{S}_{novel}$ ($C_{novel}\ge N$) and $K$ is the number of images per class in $\mathcal{S}_{novel}$. For one-shot learning, $K=1$. 
The task is to train a image classifier over $\mathcal{S}_{novel}$. Typically, we also have an additional dataset $\mathcal{D}_{base}=\{(\bm{x_i}, y_i)\}$, where $y_i \in C_{base}$, and ${C}_{base}$ $\cap$ ${C}_{novel}$ = $\emptyset$. $\mathcal{D}_{base}$ has many images for each class in $C_{base}$.

\noindent\textbf{Episodic Training}
O. Vinyals \etal \cite{matchingnet} propose an episodic paradigm for meta-learning based few-shot learning. For N-way K-shot learning, a training episode is constructed by sampling N classes from $C_{base}$ ($|C_{base}|>>N$), K images for each of these classes from $\mathcal{D}_{base}$ as the support set, and multiple query images for each of these classes from $\mathcal{D}_{base}$ as the query set. The classifier is trained over the support set and evaluated on the query set. The evaluation result is used to update the meta-learner. The idea behind this paradigm is to mimic the test setting during training, taking advantage of large amounts of labeled data in $\mathcal{D}_{base}$.

\noindent\textbf{Prototypical Network}
J. Snell \etal \cite{protonet} propose this simple yet effective model for few-shot learning. Following the episodic paradigm, it learns an embedding function $f(\bm{x})$ via ConvNet and generates a prototype of each class via Equation~\ref{eq:proto}. Then the probability for a query image from class c is calculated via Equation~\ref{eq:prob}. The embedding network is trained by feeding the probability and the ground truth label of the query image into the cross-entropy loss. During testing, Prototypical Network applies Nearest Neighbor ($\textit{NN}$) classifier for each query image, assigning it with the label of their nearest prototype.

\begin{eqnarray}
    \bm{p}_c &=& \frac{1}{|S_c|}\sum_{(\bm{x_i}, y_i)\in S_c}f(\bm{x_i})\label{eq:proto}\\
    P(y=c|\bm{x}) &=& \frac{e^{d(f(\bm{x}), \bm{p}_c)}}{\sum_{c'}e^{d(f(\bm{x}), \bm{p}_{c'})}} \label{eq:prob}
\end{eqnarray}

\begin{table*}[]
\caption{Summary of the datasets}
     \begin{center}
    \scalebox{0.95}{
        \begin{tabular}{l|cccccc}
        \hline
Datasets     & Number of images & Seen classes & Unseen classes & Resolution & Fine-grained & Strictly balanced \\
\hline\hline
\textit{mini}ImageNet  & 60,000           & 64 + 16        & 20             & Medium     & No           & Yes    \\
\hline
CIFAR-100    & 60,000           & 64 + 16        & 20             & Low        & No           & Yes               \\
\hline
Caltech-256   & 30,607           & 100 + 56       & 50             & High       & No           & No                \\
\hline
CUB-200       & 11,788           & 100 + 50       & 50             & High       & Yes          & No \\     \hline          
    \end{tabular}
    }
     \end{center}
    \label{dataset}
\end{table*}

\subsection{Learning to Restore Prototypes}
\label{ourmethod}
Even though Prototypical Network has shown to be effective for few-shot learning, its performance for one-short learning is barely satisfactory. We conjecture that the inaccurate classification is due to the learned prototypes. When there is only one image per class provided for training, the prototype is just the feature of this image, which would be far away from the class center if the image is not discriminative for this class. Consequently, the classification boundary would be shifted into a inappropriate position. In this section, we introduce our proposed model, dubbed as $\textit{RestoreNet}$, to ``restore" the prototype, i.e., moving it closer to the class center where the true prototype is more likely to situate. 

Firstly, we adapt Prototypical Network to train the feature embedding function $f(\bm{x})$. Different to the original Prototypical Network, following \cite{DEML}, we add an additional label classification branch as a regularization, which consists of a two-layer MLP network, a Softmax layer and a cross-entropy loss. The parameters of the embedding network are trained w.r.t the summation of the new loss and the original loss from Prototypical Network. Once this step is done, we freeze the parameters in $f(\cdot)$ and use it just for feature extraction. We use the same naming style in \cite{DEML} and denote this network as DEML+Prototypical Nets. This adaption is able to realize more than one percent enhancement over the original Prototypical Network. We use DEML+Prototypical Nets as the baseline in our experiments.

Secondly, we train a regression model to restore the prototypes. The regression model is a MLP network, whose output is denoted as $M(\bar{\bm{x}})$ where $\bar{\bm{x}}=f(\bm{x})$. The loss function is the squared Euclidean distance between $M(\bar{\bm{x}})$ and the true prototype of each class. The training data is collected as follows. For each class $c\in C_{base}$, we generate its prototype according to Equation~\ref{eq:proto} where $S_c$ includes all images from class c in $D_{base}$. In this way, the prototype, denoted by $\bm{t}_c$, is considered to be discriminative for the class and is thus used as the target, i.e., the truth prototype, to train $M(\cdot)$.
Next, for each target prototype $\bm{t}_c$, we select its $\lambda$ farthest images from class c, which are considered as noisy and non-discriminative examples of class c.
Each of the selected noisy images and its target prototype constitutes a training pair.

\begin{figure}
\centering
\includegraphics[width=0.47\textwidth]{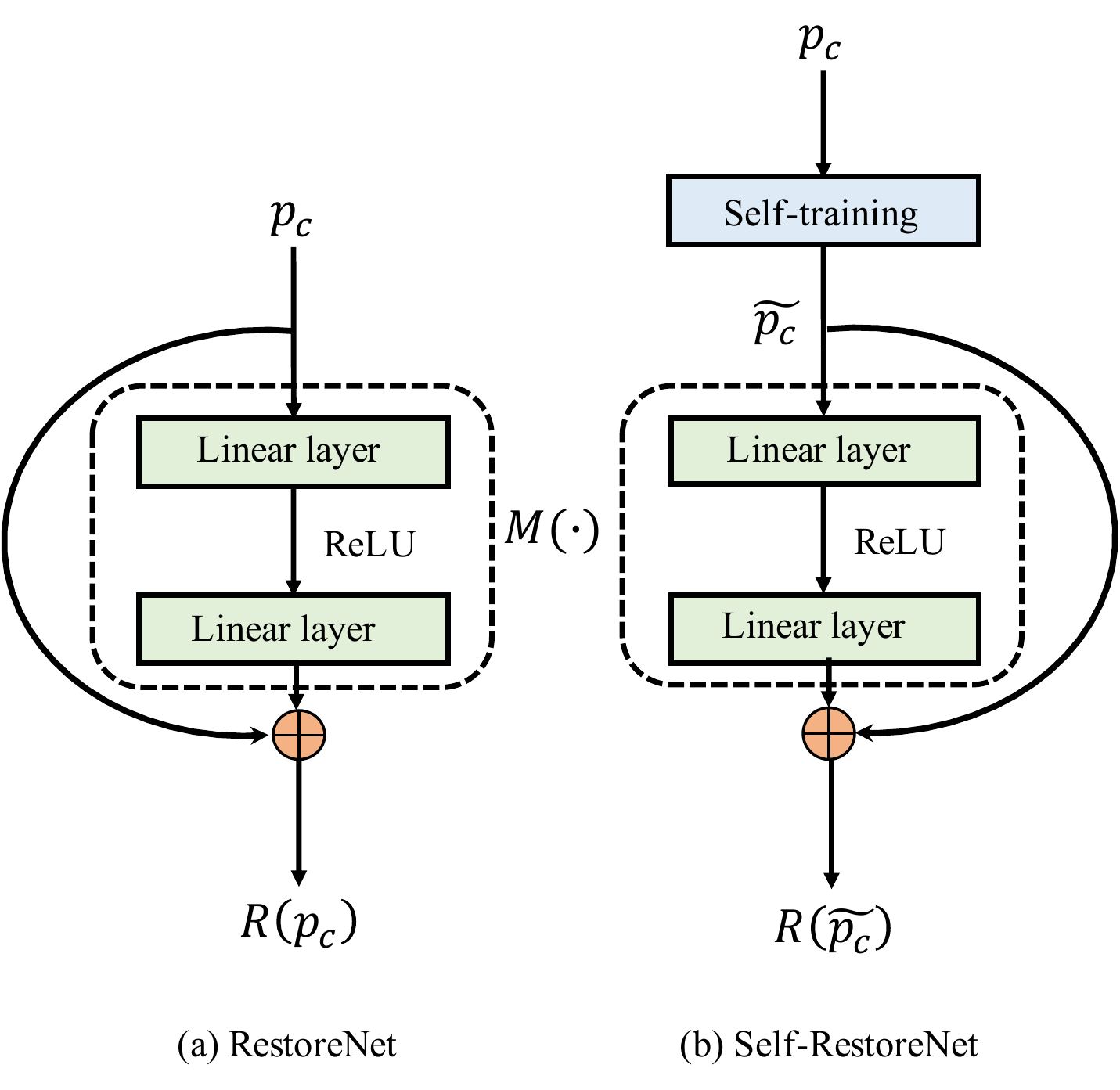}
\caption{Workflow for RestoreNet and Self-RestoreNet}
\label{workflow}
\end{figure}

Lastly, during inference, we average $M(\bm{p_c})$ and $\bm{p_c}$ as the final prototype $R(\bm{p_c})$ following Equation~\ref{eq:one}. The structure looks like that in ResNet.
We adopt this skip-connection structure for 1) ensemble modelling; 2) data augmentation by considering $M(\bm{p_c})$ as a new image; 3) avoid mis-transformation by the regression network. 
We compute the distance between $R(f(\bm{x}))$ and $f(\bm{x'})$ directly for nearest neighbor classification, where $\bm{x'}$ is the query image and $\bm{x}$ is the image from the support set $S_{novel}$. The workflow for \emph{RestoreNet} is depicted in Figure \ref{workflow}(a).
\begin{eqnarray}\label{eq:one}
    R(\bm{p_c}) = \frac{1}{2}M(\bm{p_c}) + \frac{1}{2}\bm{p_c}
\end{eqnarray}

\subsection{Self-Training}
\label{semisupervised}
In real-life, the query images are usually processed in batch to improve the system throughput\cite{rafiki}. To further improve the prototype, we try to exploit the query images in the query set via self-training~\cite{semi,selfjig}. When classifying an image, we use all other images
in the query set to constitute an unlabeled set $\textit{U}$. For each initial prototype, we retrieve its $\gamma$ nearest images from $\textit{U}$ and add them into the support set to refine the prototype (Equation~\ref{eq:proto}). We denote the refined prototype as $ \widetilde{\bm{p_c}}$. All procedures described above just happen in inference without retraining. We depict the workflow in Figure \ref{workflow}(b). Note that this self-training step is optional in our model, denoted as \emph{Self+RestoreNet}. We adopt this scheme to show that $\textit{RestoreNet}$ can be easily combined with other methods to realize further performance improvement.

\section{Experiments}

\subsection{Datasets}
We evaluated our model on multiple benchmark datasets for one-shot image classification. They are \textit{mini}ImageNet, CIFAR-100, Caltech-256 and Caltech-UCSD Birds-200-2011 (CUB-200). These datasets span a large variety of properties and can simulate various application scenarios. We adopt the same data splits with previous works \cite{deltaencoder,semanticaug,DEML,metalearnlstm}. More details about those datasets are summarized in Table \ref{dataset}.

\subsection{Implementation details}
In order to make a fair comparison with the state-of-the-art algorithms, we follow \cite{semanticaug,selfjig,SNAIL,DEML,deltaencoder} and adopt ResNet18 \cite{resnet} as our feature extractor, which outputs a 512-dimensional vector as the feature for each image. As described in Section \ref{ourmethod}, during training we add a label classifier branch to the Prototypical Network. This additional branch is implemented via a MLP whose hidden layer has 256 units and output layer has $|C_{base}|$ units. ReLU is used as the activation layer. 
The two loss terms are weighted summed as the total loss for the whole network. Note that the additional image classifier will be discarded after training.
We use 30-way 1-shot episodes with 10 query images per category to train the network. In each epoch,
600 such episodes are randomly sampled. The learned feature extractor is applied in all subsequent experiments. We optimize the networks via Adam with a initial learning rate $10^{-3}$, annealed by half for every 20 epochs.

$\textit{RestoreNet}$ is trained after we obtain the feature extractor. We use a two-layer MLP for $M(\cdot)$, where the hidden layer has 256 units. ReLU is chosen as the activation function for the hidden layer. The output layer has the same number of units as the input. We tune $\lambda$, which is the number of selected noisy images per class, on the validation dataset. It is 100, 30, 5 ,1 for \textit{mini}ImageNet, CIFAR-100, Caltech-256 and CUB-200 respectively. Adam with a fixed learning rate $10^{-3}$ is used to train the regression netowrk, i.e., $M(\cdot)$. 
Training $\textit{RestoreNet}$ is fast, which takes only tens of seconds.
We report the average performance on 10,000 randomly generated episodes from the test split. Each episode contains 30 query images per category.

For Self-training scheme, as described in Section \ref{semisupervised}, we take the whole query set (exclude current image to be classified) as the unlabeled set \textit{U} in experiments on CIFAR-100, Caltech-256 and CUB-200. While in experiments on \textit{mini}ImageNet, we adopt an alternative implementation method following \cite{selfjig} for fair comparison with it. Instead of applying self-training over the query set, we supply another unlabeled images set $\textit{U}_\textit{novel}$ to each episode as the unlabeled set \textit{U}. 
For an episode, its supplied unlabeled images set $\textit{U}_\textit{novel}$ has the same samples distribution, i.e., number of examples per class, with the query set. 
We stipulate that $\gamma$ = 4, i.e., each initial prototype retrieves its 4 nearest images from unlabeled set \textit{U} to do self-training.

\subsection{Results and discussion}
We report the performance of our model on miniImageNet in Table \ref{miniimagenet}. It can be found that our proposed method outperforms the state-of-the-art methods in one-shot learning.
For fair comparison, we reimplement Prototypical Network\cite{protonet} by replacing its feature extractor  with ResNet18\cite{resnet}. Existing papers have done the experiments of Matching Network\cite{matchingnet}, Relation Network\cite{relationnet} and MAML\cite{maml} using ResNet18 as the feature extractor. Therefore we directly take the corresponding results from the paper~\cite{semanticaug}. Note that some approaches in Table~\ref{miniimagenet} require additional data, \eg DEML+Meta-SGD\cite{metasgd} needs an external large-scale dataset \textit{ImageNet-200}\cite{metasgd}; 
Dual TriNet\cite{semanticaug} uses word embeddings or human-annotated class attributes for data augmentation; Delta-encoder\cite{deltaencoder} needs to be exposed to large number of samples to extract intra-class deformations; 
Self-training based methods like Self-Jig\cite{selfjig} and Self-RestoreNet also require access to external samples. Our method, RestoreNet (not Self-RestoreNet), can work without any restriction. What is more, compared with other models, ours is significantly simpler in structure.

\begin{table}[]
  \caption{The 5-way, 1-shot classification results(\%) on \textit{mini}ImageNet. The ``$\pm$'' indicates 95\%
    confidence intervals over tasks. The ``$\pm$'' is not reported in Delta-encoder.}
\begin{center}
   \scalebox{0.95}{
    \begin{tabular}{lc}
\hline
Models            & 1-shot Acc.  \\
\hline\hline
Meta-LSTM \cite{metalearnlstm} & 43.44$\pm$0.77 \\
\hline
Meta-Net \cite{metanetworks}          & 49.21$\pm$0.96 \\
\hline
Matching Nets \cite{matchingnet}     & 43.56$\pm$0.84\\
(Deep)       & 47.89$\pm$0.86\\
                  \hline
ProtoNets \cite{protonet}     & 49.42$\pm$0.78\\
(Deep)                  & 56.35$\pm$0.77\\ \hline
Relation Nets \cite{relationnet}     & 50.44$\pm$0.82\\
 (Deep)                 & 57.02$\pm$0.92\\
                  \hline
MAML \cite{maml}              & 48.70$\pm$1.84\\
  (Deep)                & 52.23$\pm$1.24\\
\hline
Meta-SGD \cite{metasgd}          & 50.47$\pm$1.87\\
 (Deep)                 & 52.31$\pm$1.14 \\
\hline
SNAIL \cite{SNAIL}             & 55.71$\pm$0.99  \\
\hline
DEML+Meta-SGD \cite{DEML}     & 58.49$\pm$0.91 \\
\hline
Dual TriNet \cite{semanticaug}&58.12$\pm$1.37 \\
\hline
Delta-encoder \cite{deltaencoder}     & \textbf{59.90 }      \\
\hline
Self-Jig \cite{selfjig}          & 58.80$\pm$1.36 \\
\hline\hline
RestroreNet (Ours)     & \textbf{59.28$\pm$0.20}       \\
\hline
Self-RestroreNet (Ours)          & \textbf{61.14$\pm$0.22}  \\
\hline
    \end{tabular}
    }
       \end{center}
    \label{miniimagenet}
\end{table}

$\textit{RestoreNet}$ is also able to achieve state-of-the-art or even superior performance on the other three datasets, CIFAR-100, Caltech-256 and CUB-200, as presented in Table \ref{other3}. We infer that our approach 
can take effects under different tasks and scenarios.

\begin{table*}[]
\caption{The 5-way, 1-shot classification results(\%) on CIFAR-100, Caltech-256 and CUB-200. The ``$\pm$'' indicates 95\%
    confidence intervals over tasks. Note that ``$\pm$'' is not reported in some previous works. Average performances on 600 randomly generated episodes are reported.}
    \begin{center}
    \begin{tabular}{l|ccc}
    \hline
      Models   & CIFAR-100 & Caltech-256 & CUB-200  \\
      \hline\hline
      Matching Nets \cite{matchingnet}  & 50.53$\pm$0.87  & 48.09$\pm$0.83 &49.34 \\
      \hline
      MAML \cite{maml} & 49.28$\pm$0.90 & 45.59$\pm$0.77  &38.43 \\
      \hline
      DEML+Meta-SGD \cite{DEML} & 61.62$\pm$1.01 & 62.25$\pm$1.00 &66.95$\pm$1.06 \\
      \hline
      Dual TriNet \cite{semanticaug}&63.41$\pm$0.64 & 63.77$\pm$0.62 & 69.61$\pm$0.46 \\
      \hline
      Delta-encoder \cite{deltaencoder}& 66.7 &\textbf{73.2}  & 69.8 \\
      \hline\hline
      RestroreNet (Ours) & \textbf{66.87$\pm$0.94} & 64.10$\pm$0.89 & \textbf{74.32$\pm$0.91} \\
      \hline
      Self+RestroreNet (Ours) & \textbf{69.09$\pm$0.97} & \textbf{68.28$\pm$0.96} & \textbf{76.85$\pm$0.95} \\
      \hline
    \end{tabular}
    \end{center}
    \label{other3}
\end{table*}

\subsection{Ablation study}

\textbf{Does $\textit{RestoreNet}$ work consistently?} To find out if $\textit{RestoreNet}$ takes effect across different tasks and how much enhancement it is able to achieve, we strictly control variables and conduct a series of N-way one-shot experiments on \textit{mini}ImageNet. The results averaged over 600 randomly generated test episodes are presented in Table \ref{nway_t}. We can find that $\textit{RestoreNet}$ achieves obvious and consistent improvement on all these tasks. Baseline and RestoreNet use $\bm{p_c}$ and $R(\bm{p_c})$ as the prototypes respectively (Figure \ref{workflow}(a)).

\begin{table*}[]
\caption{N-way one-shot tasks results(\%) with the enhancements of RestoreNet on \textit{mini}ImageNet. The ``$\pm$'' indicates 95\%
    confidence intervals over tasks. Average performances on 600 randomly generated episodes are reported.}
    \begin{center}
    \scalebox{0.95}{
    \begin{tabular}{l|ccccccc}
    \hline
Models                            & 5-way & 7-way & 9-way & 11-way & 13-way & 15-way & 20-way \\
\hline\hline
Baseline & 57.67$\pm$0.83 & 49.11$\pm$0.68 & 43.38$\pm$0.54  & 38.92$\pm$0.44  & 35.29$\pm$0.40  & 32.48$\pm$0.35  & 27.69$\pm$0.27  \\
\hline
RestroreNet & 59.56$\pm$0.84 & 50.55$\pm$0.68 & 44.54$\pm$0.55 & 39.98$\pm$0.43   & 36.34$\pm$0.39  & 33.52$\pm$0.35  & 28.48$\pm$0.27\\
\hline\hline
Enhancement&	1.89&	1.44&	1.16&	1.06&	1.05&	1.04&0.79\\
\hline
    \end{tabular}
    }
    \end{center}
    \label{nway_t}
\end{table*}

\begin{table}[]
\caption{Five-way one-shot classification accuracy(\%) on miniImageNet. Results presented are for $\bm{p_c}$, $\widetilde{\bm{p_c}}$ and $R({\widetilde{\bm{p_c}}})$ respectively (shown in Figure \ref{workflow}(b)). The ``$\pm$'' indicates 95\%
    confidence intervals over tasks.}
    \begin{center}
    \scalebox{0.95}{
    \begin{tabular}{lc}
    \hline
         Models                             & 5-way 1-shot Acc. \\
         \hline\hline
DEML+Prototypical Nets (Baseline)  & 57.63$\pm$0.20             \\
\hline
Self training + Baseline & 59.78$\pm$0.22             \\
\hline
Self+RestroreNet                          & 61.14$\pm$0.22         \\
\hline\hline
Enhancement & 1.36\\
\hline
    \end{tabular}}
    \end{center}
    \label{selfr}
\end{table}

In order to explore whether $\textit{RestoreNet}$ is able to achieve further performance improvement when combined with other state-of-the-art algorithms, we adopt the Self-training scheme and test $\textit{RestoreNet}$ on it. As shown in Table \ref{selfr}, Self-training is an effective algorithm which can significantly boost the performance of the baseline
by more than two percent, reaching a challenging results at 59.78\%. Despite this, $\textit{RestoreNet}$ still improves the performance of the model by a obvious margin, around 1.36 percent.

\noindent\textbf{How to configure $\lambda$} During the training of \emph{RestoreNet}, only images with serious noise in each class are fed to the network. 
We select the training images in this way because RestoreNet is proposed to correct the prototype generated from the noisy image. If we use all available images in $D_{base}$, then both noisy images and normal images are included in the training data, which would confuse the model on how to restore the prototype. In other words, it would pose difficulty on the training (optimization) process. Instead, if we only select the $\lambda$ farthest images, we are likely to exclude those normal images.

To verify our assumption above, we conduct experiments on \textit{mini}ImageNet. We gradually increase $\lambda$, which is the number of noisy images collected from each category, from 100 to 600 (all images belonging to the class) and train the corresponding $M(\cdot)$. We report the 5-way 1-shot results together with the enhancements in Table \ref{trans_n}. It can be found that $\textit{RestoreNet}$ realize the best performance when $\lambda$ equals to 100. As $\lambda$ increases, in general, the enhancement effects of $\textit{RestoreNet}$ drops. And when $\lambda$ equal to 600, which is the case trying to learn an universal model, it gives us the worst performance. 

\noindent\textbf{How simple can $\textit{RestoreNet}$ be?}
Our training strategy that only feed the network with noisy images significantly reduce the difficulty of learning for \emph{RestoreNet}. It is for this reason that \emph{RestoreNet} is able to achieve good performance with a simple structure. How simple can \emph{RestoreNet} be while keeping the enhancement? To answer this question,
we reimplement the regression network for $M(\cdot)$ as the simplest network which is able to handle a 512-dimension to 512-dimension regression task. This model is just a two-layer MLP with the single hidden layer of only one units. $\lambda$ is set as 100 here. The performance of this simplest $\textit{RestoreNet}$ on \textit{mini}ImageNet is presented in Table \ref{simplest}. From the results, we can still observe enhancements, although they are slightly weaker than that of a more complex structure.

\begin{table}[]
\caption{\textit{mini}ImageNet five-way one-shot classification results(\%) for the \textbf{simplest} $\textit{RestoreNet}$(512-1-512). 
The ``$\pm$'' indicates 95\%
    confidence intervals over tasks.}
    \begin{center}
    \scalebox{1}{
    \begin{tabular}{lc}
    \hline
            & 5-way 1-shot Acc.\\
            
            \hline\hline
Baseline     & 57.67$\pm$0.83   \\  
RestoreNet  &58.98$\pm$0.83\\
Enhancement&1.31\\
\hline\hline
Self+Baseline     & 60.37$\pm$0.90\\
Self+RestoreNet &61.64$\pm$0.90\\                            
Enhancement &1.27  \\                            
\hline
    \end{tabular}
    }
    \end{center}
    \label{simplest}
\end{table}

\begin{table*}[]
\caption{\textit{mini}ImageNet five-way one-shot classification results(\%) for $\textit{RestoreNets}$ trained under different $\lambda$, which is the number of noisy samples collected from each category. Average performances on the same 10,000 randomly generated episodes are reported. The ``$\pm$'' indicates 95\%
    confidence intervals over tasks.}
     \centering
    \begin{tabular}{lcccccc}
    \hline
\textbf{$\lambda$}      & 100   & 200   & 300   & 400   & 500   & 600   \\
\hline\hline
Baseline      & 57.63$\pm$0.20 & 57.63$\pm$0.20 & 57.63$\pm$0.20 &57.63$\pm$0.20 & 57.63$\pm$0.20 & 57.63$\pm$0.20 \\
RestroreNet          & 59.28$\pm$0.20 & 58.93$\pm$0.20 & 59.12$\pm$0.20 & 58.84$\pm$0.20 & 58.55$\pm$0.20 & 58.21$\pm$0.20 \\

Enhancement   & \textbf{1.65}  & 1.30  & 1.49  & 1.21  & 0.92  & 0.58  \\
\hline\hline
Self+Baseline & 59.78$\pm$0.22 & 59.78$\pm$0.22 & 59.78$\pm$0.22 & 59.78$\pm$0.22 & 59.78$\pm$0.22 & 59.78$\pm$0.22 \\
Self+RestroreNet     & 61.14$\pm$0.22 & 61.08$\pm$0.22 & 60.68$\pm$0.22 & 60.90$\pm$0.22 & 60.41$\pm$0.22 & 60.18$\pm$0.22 \\

Enhancement   & \textbf{1.36}  & 1.30  & 0.90  & 1.12  & 0.63  & 0.40   \\
\hline
    \end{tabular}
    \label{trans_n}
\end{table*}

\subsection{Visualization}
\begin{figure}
    \centering
    \includegraphics[width=0.35\textwidth]{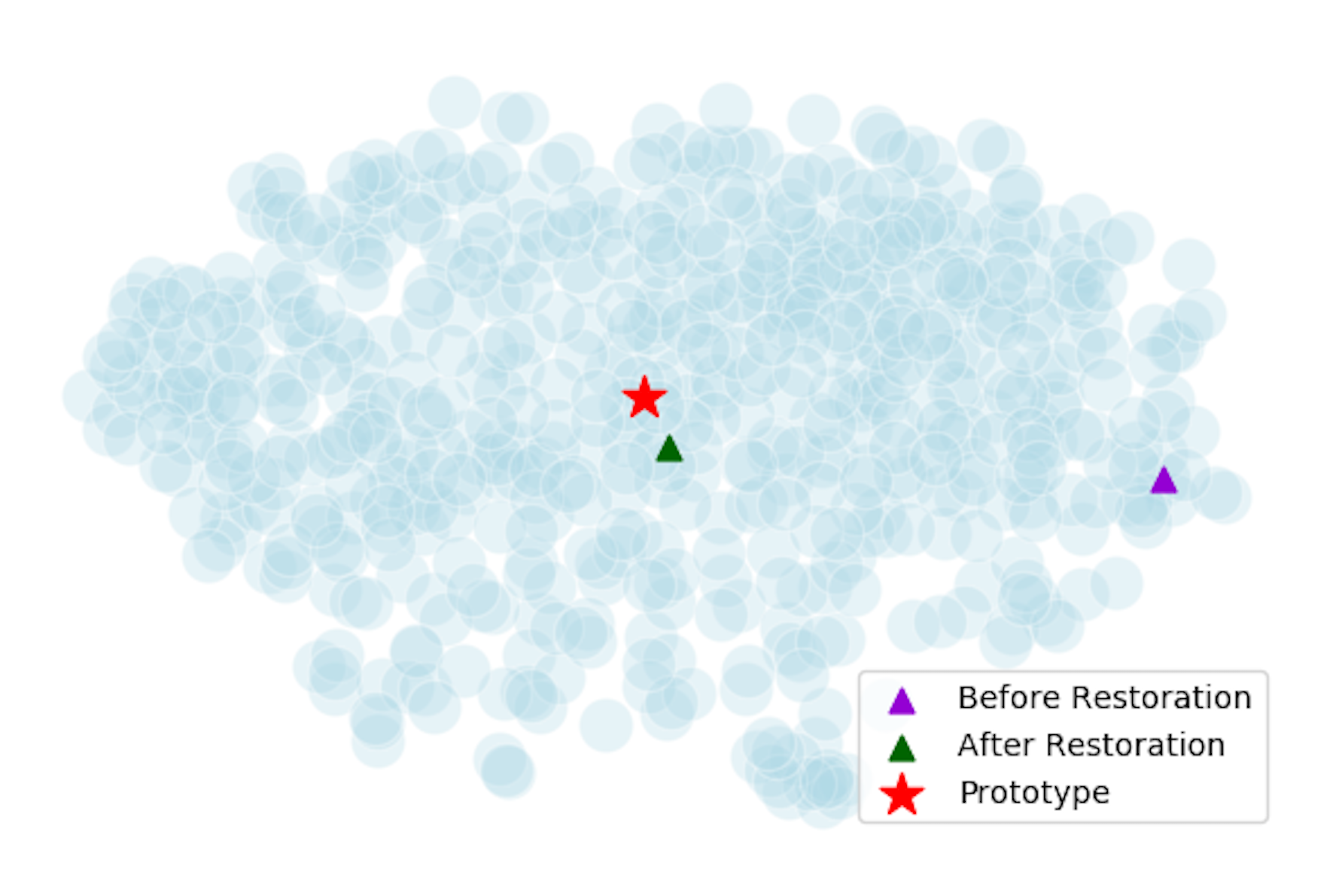}
    \caption{Visualization for images of a class, Golden Retriever, in \textit{mini}ImageNet. The red star shows the target prototype while triangles mark a proposed prototype(learned from a one-shot example) before(violet) and after(green) restoration. $\textit{RestoreNet}$ manages to move the initial proposed prototype much more close to its destination. The figure is plotted by applying t-SNE to ResNet18 features of images. Best viewed in color}
    \label{1_cluster}
\end{figure}

To understand how $\textit{RestoreNet}$ works, we select a class, Golden Retriever, from the test split of \textit{mini}ImageNet and visualize samples in this category by applying t-SNE\cite{tsne} to their feature vectors. As shown in Figure \ref{1_cluster}, the red star marks the ``true'' prototype (calculated by averaging feature vectors of all 600 images) of this class while the violet triangle represents the initial prototype. The restored prototype is marked by the green triangle. It can be found that the initial prototype (violet triangle) is moved much closer to the ``true'' prototype by the \emph{RestoreNet}.

\begin{figure}
    \centering
    \includegraphics[width=0.25\textwidth]{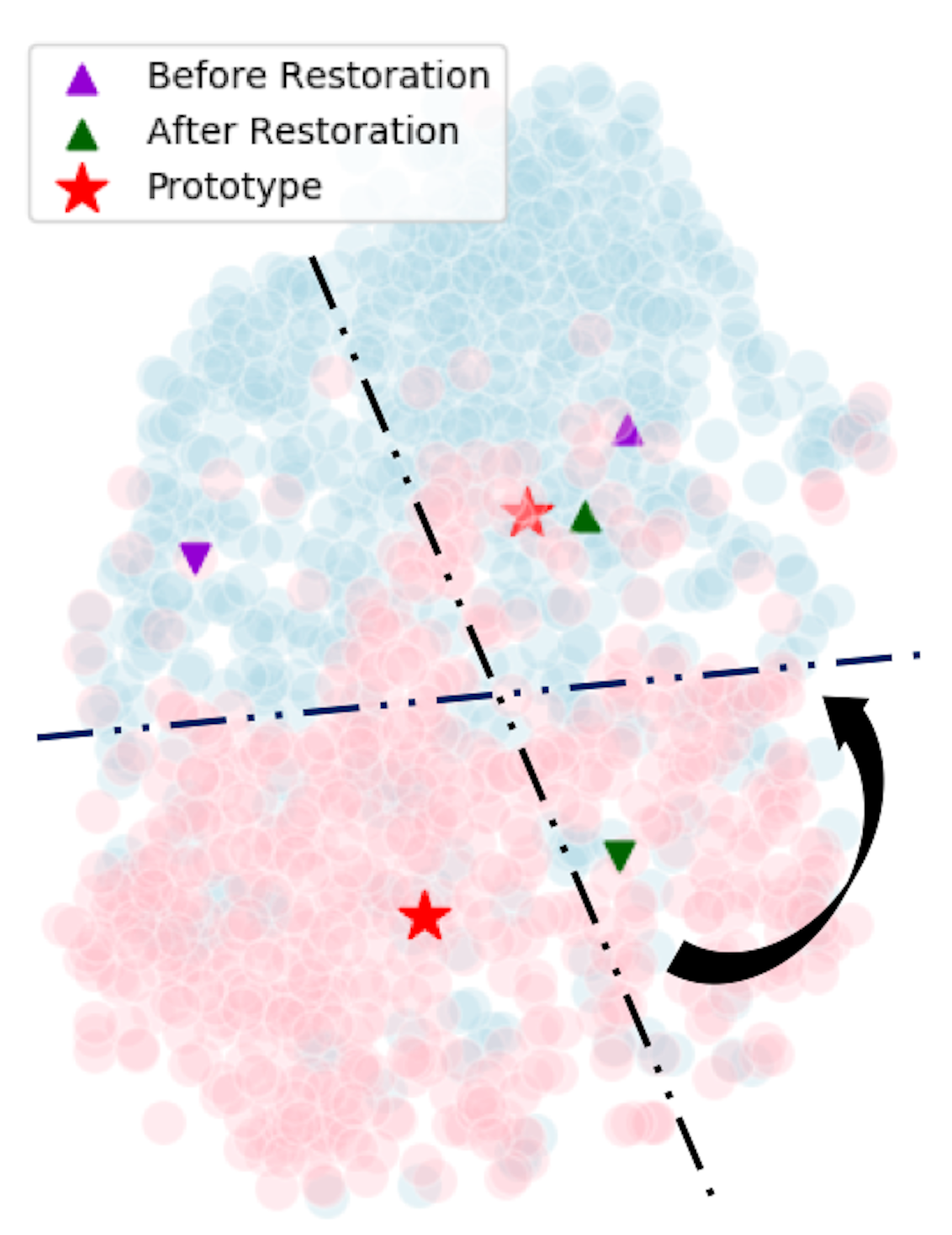}
    \caption{Visualization of a two-way one-shot classification task sampled from \textit{mini}ImageNet. After restoration, violet triangles which represent initial proposed prototypes are pushed to their corresponding green triangles. The green triangles are significantly closer to the target prototypes(marked by the red stars) compared to the violet. The figure is plotted by applying t-SNE to ResNet18 features of images. Best viewed in color}
    \label{2_cluster}
\end{figure}

We further visualize a two-way one-shot classification task which is sampled from \textit{mini}ImageNet. The two classes to be classified are Golden Retriever (blue) and Alaskan Malamute (pink). As shown in Figure \ref{2_cluster}, violet triangles mark samples in the support set (one image per class) while points (blue and pink) represent images in the query set. After applying \emph{RestoreNet}, violet triangles which represent the initial prototypes are pushed to their corresponding green triangles. It can be found that the green triangles are closer to the target true prototypes (marked by the red stars) compared to the violet. And if we plot the perpendicular bisector of two triangles in the same color, it will give us a decision boundary of the two class. It is obvious that decision boundary produced by the green triangle pair is much better than that produced by the violet ones.

\subsection{Statistical analysis}
To find out if \textit{RestoreNet} works as expected, i.e., moving feature vectors of samples more close to their clusters' centers, we calculate the average distances between 
the learned prototype (from each test image) and the corresponding class center. We compare three types of prototypes, namely the output from the feature extraction network ($p$), the direct output from \textit{RestoreNet} $M(p)$ and the output after applying skip-connection $R(p)$. Equation \ref{eq:one} illustrates the relation among $p$, $M(p)$ and $R(p)$. As shown in Table \ref{dist}, the statistics demonstrate that \textit{RestoreNet} takes effect and the average distances using $M(p)$ and $R(p)$ are reduced compared with the distance calculated using $p$.

\begin{table}[]
\caption{Distances between samples and their corresponding class centers, averaged among all samples in the test set.}
 \centering
\begin{tabular}{c|ccc}\hline
 Average dist. (test set)       & $p$ & $M(p)$ & $R(p)$ \\
                  \hline\hline
\textit{mini}ImageNet  & 1.7981         &  0.9386          &  1.0822       \\
\hline
                  
CIFAR-100  & 1.7791         &  1.3790          &  1.1305       \\
\hline
Caltech-256  & 1.5930         &  1.2339          &  1.0101       \\
\hline
CUB-200  & 1.5142         &  1.3592          &  1.0403       \\
\hline
\end{tabular}
\label{dist}
\end{table}

\section{Conclusion}
For one-shot learning, if the training image is far away from the class center, then metric-learning-based approaches would fail to find a good decision boundary in the feature space. In this work, we propose a simple yet effective model (\emph{RestoreNet}) to move the the class prototype constructed from the (noisy) image closer to the class center. Experiments demonstrate that our model obtains superior performance over the state-of-art approaches on a broad range of tasks. In addition, we conduct ablation study and visualize the prototypes to verify the effects of \emph{RestoreNet}.

\section*{Acknowledgements}
This work is supported by the National Research Foundation, Prime Ministers Office, Singapore under its National Cybersecurity
RD Programme (No. NRF2016NCR-NCR002-020), and FY2017 SUG Grant. We would like to thank all anonymous reviewers for their valuable comments.

\bibliography{6350_ref.bib}
\bibliographystyle{aaai}
\end{document}